# CVFC: Attention-Based Cross-View Feature Consistency for Weakly Supervised Semantic Segmentation of Pathology Images

Liangrui Pan
College of Computer Science and
Electronic Engineering
HunanUniversity
Chang Sha, China
panlr@ hnu.edu.cn

Lian Wang
College of Computer Science and
Electronic Engineering
HunanUniversity
Chang Sha, China
lianwang@hnu.edu.cn

Zhichao Feng
Department of Radiology
Third Xiangya Hospital
Central South University
Chang Sha, China
fengzc2016@163.com

Liwen Xu*
College of Computer Science and
Electronic Engineering
HunanUniversity
Chang Sha, China
xuliwen@hnu.edu.cn

Shaoliang Peng*
College of Computer Science and
Electronic Engineering
Hunan University
Chang Sha, China
slpeng@hnu.edu.cn

*Abstract*—Histopathology image segmentation is the gold standard for diagnosing cancer, and can indicate cancer prognosis. However, histopathology image segmentation requires high-quality masks, so many studies now use image-level labels to achieve pixel-level segmentation to reduce the need for fine-grained annotation. To solve this problem, we propose an attention-based cross-view feature consistency end-to-end pseudo-mask generation framework named CVFC based on the attention mechanism. Specifically, CVFC is a three-branch joint framework composed of two Resnet38 and one Resnet50, and the independent branch multi-scale integrated feature map to generate a class activation map (CAM); in each branch, through down-sampling and The expansion method adjusts the size of the CAM; the middle branch projects the feature matrix to the query and key feature spaces, and generates a feature space perception matrix through the connection layer and inner product to adjust and refine the CAM of each branch; finally, through the feature consistency loss and feature cross loss to optimize the parameters of CVFC in co-training mode. After a large number of experiments, An IoU of 0.7122 and a fwIoU of 0.7018 are obtained on the WSSS4LUAD dataset, which outperforms HistoSegNet, SEAM, C-CAM, WSSS-Tissue, and OEEM, respectively.

## I. INTRODUCTION

With the development of artificial intelligence, computational pathology allows pathologists to gradually reduce the heavy workload, such as image annotation, classification, segmentation, etc [1]–[4]. In computational pathology, the segmentation of histopathological images has been a hotspot of extensive research. It is closely related to the analysis of the tumor microenvironment. Tumor microenvironment analysis plays a vital role in tumor occurrence and development and affects the treatment effect and prognosis of patients [5]. Accurate segmentation of tumor histopathology can also guide doctors' medication decisions and clinical treatment plans.

To achieve high histopathology segmentation accuracy, most pathology images require mask-level labels for supervised learning [6]. Different tissues, cells, and structures in histopathological images are quite different. In order to obtain high-quality pathology image masks, pathologists are required to label them manually. This process is time-consuming and has uncertain errors in the labels of different pathologists. Therefore, sufficient pathology physicians are also a challenge in the development of computational pathology.

In order to reduce the annotation time of pathologists, coarse label annotation of images can help them reduce most of their working time. Currently, weakly supervised semantic segmentation based on image label level is mainly based on class activation map (CAM) methods [7]. The image is passed to the convolutional neural network (CNN) through forward propagation to obtain the last layer of feature maps, and then average pooling is performed on the last layer of feature maps. Convolution or entire connection operations map the features to the corresponding category weight vectors. The weights are the same as the last layer feature maps are multiplied by channels to obtain class activation maps. However, an essential disadvantage of CAM is that it cannot refine the boundary of objects, which leads to uncertainty in the local area of the image boundary. This will bring troubles to pathology and doctors' diagnosis and guidance.

Inspired by CAM, we need to solve the problem of boundary region uncertainty in practical applications. Therefore, attention-based methods are proposed to map image features in query and key spaces [8]. Second, we can obtain the feature-space attention-aware matrix to refine the CAM to solve the problem of CAM boundary ambiguity. Assuming that there is only one standard mask for a known graph, that independent branch can theoretically generate three masks, and these three masks are the same. Based on this idea, we propose a cross-view feature consistency principle to constrain the results of CAM generated on independent branches, and optimize the parameters of CVFC through co-training.

Therefore, the main innovations of this paper can be summarized as follows:

(1) We propose a CVFC framework for WSSS of pathological images. It is constructed by a three-branch

combined CNN network, and multi-scale integrated feature maps are used to generate CAM in an adaptive manner.

(2) The principle of attention-based cross-view feature consistency adjusts and refines the CAM of each branch through a spatial attention matrix.

(3) Through extensive experiments, CVFC framework achieves the best segmentation performance on WSSS4LUAD datasets.

## II. RELATED WORKS

### A. WSSS

Weakly supervised semantic segmentation is an approach to semantic segmentation with minimal labeling information. In order to solve the difficult and costly problem of labeling data in traditional semantic segmentation. Some weakly supervised semantic segmentation methods address this problem by utilizing data that only has image-level labels. For example, CAM is an attentional mechanism based on global average pooling that enables predictions from image level to pixel level [9]. On the PASCAL VOC 2012 dataset, AffinityNet implements semantic propagation by random wandering to predict the semantic affinity between a pair of neighboring image coordinates, relying only on image-level class labels for optimal segmentation [10]. ECS-Net generates pixel-level labels by predicting the segmentation results of the image processed by the CAM technique [11]. ReCAM first extracts the feature pixels of each class using CAM and uses them along with the class labels to learn softmax cross entropy loss (SCE) another fully connected layer (after the backbone) that converges, i.e., yields a high-quality mask [12].C^2AM uses a new contrast loss to force the network to decouple foreground and background using a class-independent activation mapping [13]. The class-independent activation mapping learned by our method generates more complete object regions when the network is guided to distinguish between foreground and background across images; to eliminate intra-class inconsistencies caused by local contextual variance, Sun-Ao et al. further proposed partial class activation attention (PCAA) that simultaneously utilizes both local and global class-level representations for attention computation [14].

Recent research work shows that Transformer-based Token Contrast (CoTo) [15] includes a Patch Token Contrast Module (PTC) that supervises the final patch tokens using pseudo-token relations derived from intermediate layers, allowing them to align semantically area, resulting in a more accurate CAM. Second, to further distinguish low-confidence regions in CAM, they design a token-like contrast module (CTC). Secondly, Rongtao et al. proposed a Wave-like Class Activation Map (WaveCAM) in terms of representation fusion and dynamic aggregation of representations [16]. It includes foreground-aware representation modeling that enhances perception of foreground information, and foreground-independent representation modeling that enhances perception of foreground-independent information, as well as a representation-adaptive fusion module that fuses the two representations. Finally, Zicheng et al. propose a new conflict-based cross-view consistency (CCVC) method based on the dual-branch joint training framework, which aims to force two subnetworks to learn informative features from unrelated views [17].

### B. WSSS for Medical Image

To free pathologists from the heavy annotation burden, many AI-based WSSS methods have been proposed in recent years. Jun et al. proposed a joint fully convolutional and graph convolutional network (FGNet) for weakly supervised learning of pathological images and realized automatic segmentation [18]. The feature extraction module and classification module in FGNet are connected by dynamic superpixel operations, which enables joint training and reduces inaccurate image inference by constraints of different certainty ranges [18]. Julio et al. proposed WeGleNet, a weakly supervised training convolutional neural network, based on multi-class segmentation layers, global aggregation and background class slices after the feature extraction module, for model loss estimation during training [19]. Kailu et al. proposed a novel weakly supervised framework for SA-MIL, which introduces a self-attention mechanism to capture the global correlation among all instances [20]. Second, the method uses contextual information to make up for the shortcomings of independent examples in media and information literacy. Han et al.'s CAM-based model generates masks via image-level labels and achieves tissue-semantic segmentation via multi-layer pseudo-segmentation [21]. These methods shed light on weakly supervised segmentation of histopathology images.

## III. METHODOLOGY

This paper proposes a feature-consistency end-to-end label generation framework named CVFC based on the attention mechanism. As shown in Fig. 1, CVFC is mainly a three-branch framework composed of two Resnet38 and one Resnet50 [22], [23]. Each branch first extracts the features of the training image and then performs global average pooling (GAP) on the feature map to generate a CAM [24]. The CAMs of branch 1 and branch 3 will be down-sampled to a suitable shape and then expanded in the feature space. The image features of branch 2 integrate the features extracted from the training image and Resnet38 and then carry out the spatial expansion of the features. Image features flatten the features along the spatial dimension and then apply two fully connected layers to map the features to the query and key spaces. The softmax of the inner product of the query and key space matrices calculates the space-aware attention matrix [25]. Then the matrix is multiplied by the feature matrix in branch one and branch three and mapped to the attention feature space. We use the refined CAM of branch 1 and branch 3 to constrain the segmentation effect of branch 2 by feature consistency across views. A refined pseudo-label is finally generated by the multiplication of the CAM provided by branch two and the space-aware attention matrix.

### A. The Three-Branched Backbone

Resnet is an important feature extractor in CVFC, as shown in Fig. 1. Resnet is proposed mainly to solve the problem of gradient vanishing in deep neural networks. It mainly adopts the method of residual learning, which introduces residual connectivity in the network so that the network can learn the residual mapping instead of learning the underlying mapping relations, which can make the network design 'deep'. Therefore, Resnet builds a number of residual modules, each of composed of multiple convolutional layers, normalization layers, and ReLu activation functions, which can to capture more semantic features at different depths of the network [26]. Multiple residual modules are combined together to form a deep convolutional neural network capable of learning a hierarchical representation of the input, and each residual fast mitigates the problem of gradient vanishing by jumping connections to make the gradient flow more easily during backpropagation.

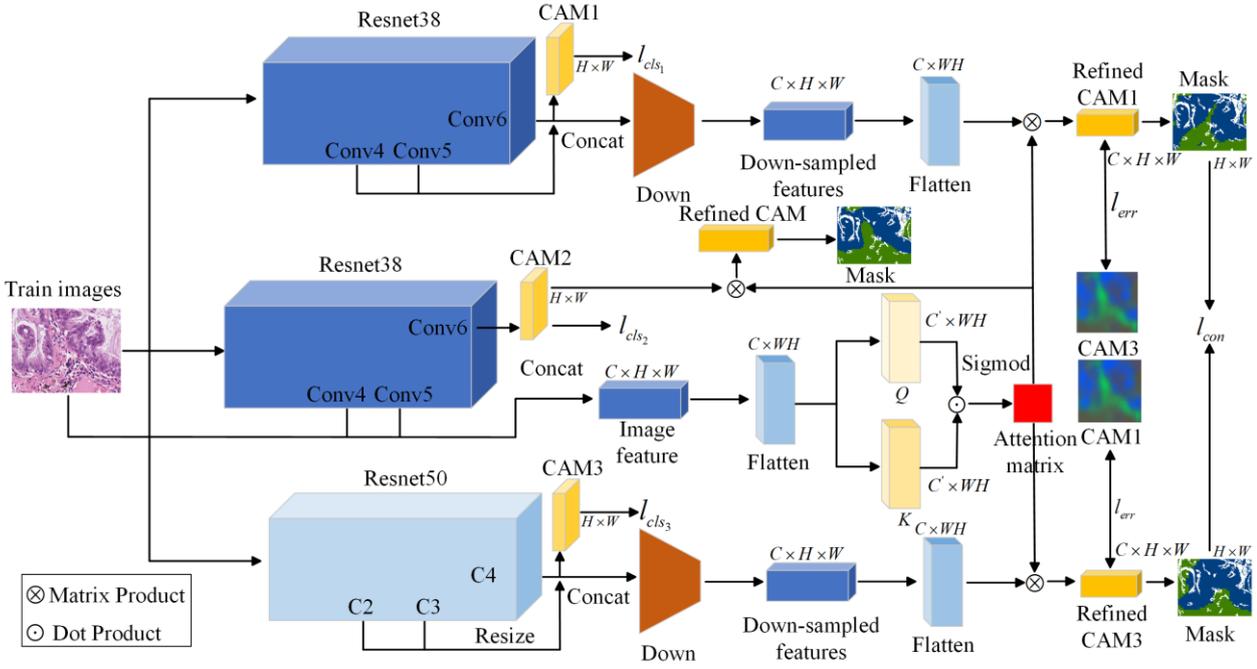

Fig. 1. Block diagram of the CVFC process.

Our proposed CVFC is composed of three common Resnet, branch one and branch three are composed of Resnet38, and branch two is composed of Resnet50 [22], [23]. Resnet38 is composed of a 38-layer network. Resnet50 is composed of a 50-layer network. We save the feature maps of different layers in the three branches to maximize the semantic information of the training images.

*B. Adaptive CAM Generation*

Before introducing adaptive CAM generation, we review the generation of CAM. CAM mainly generates CAM by performing global average pooling operations on features. The image features are first obtained in Resnet. After that, GAP is used to reduce the spatial resolution, and then the output is passed to the fully connected layer to obtain the probability score for classification. To generate the CAM, we need to compute a weighted sum based on the weights of the fully connected layer. Obviously, generating the CAM in an end-to-end manner is not feasible because it requires additional operations after forward propagation. Note that the global average pooling and the fully connected layer in the CAM generation process can be regarded as linear operations, and the fully connected layer can be equivalent to a $1\times 1$ convolutional layer. In order to simplify the CAM generation process, a strategy to rewrite it into a one-step operation is proposed:

$$\begin{aligned} A_{cam} &= Conv_{1\times 1}(X_{in}), \\ S_{cam} &= Sigmoid(AvgPool^{1\times 1}(A_{cam})) \end{aligned} \quad (1)$$

Where $A_{cam} \in R^{C\times H\times W}$ is the activation mapping graph, $S_{cam} \in R^{C\times 1\times 1}$ is the class classification score. $C$ represents the number of labels in the class activation graph, and $AvgPool^{1\times 1}$ describes the process of global average pooling and generates an $1\times 1$-sized output.

Given that this weakly supervised segmentation task provides only image-level labeling. The global average pooling layer becomes an important tool for generating CAM. However, it ignores the spatial relationships of different classes of features. Therefore, we replace the GAP operation with adaptive average pooling, which automatically adapts to produce a fixed-size output based on the input size. Unlike traditional global average pooling, the traditional global average pooling operation divides the entire feature map into fixed-size blocks and averages the elements within each block. With adaptive global average pooling, we get a fixed-dimension feature representation regardless of the variation in the input feature map size. This helps to reduce the dimensionality of the feature map, extract more global and abstract feature information, and can be applied to input images of different sizes.

*C. Attention-Based Cross-View Consistency of Features*

To illustrate our newly proposed attention-based approach for cross-view consistency, we use three branching networks based on co-training, i.e., for $\psi_1, \psi_2, \psi_3$. They have similar architectures but, do not share the parameters of the sub-networks. Recall that our goal is to make the three sub-networks produce similar results when reasoning about the mask of the same image. Therefore, the masks predicted by the sub-networks would need to be feature consistent across views on different branches.

Branch networks $\psi_1$ and $\psi_3$ obtain CAMs through preliminary feature extraction, where branch network $\psi_1$ integrates feature maps from layers Conv4, Conv5, and Conv6 and then generates pseudo-masks, and the CAMs generated from the features after layer Conv6 are optimized for the classification task. The branch network b integrates the feature maps from layers C2, C3 and C4 and then generates pseudo-masks. the CAMs generated from the features after layer C4 are also optimized for the classification task. To ensure that the sizes of CAMs generated from different networks remain consistent, we process features using down-sampling and project these feature size expansions onto the space of attention. The branching network $\psi_2$ integrates the feature maps from the training data $X_{images}$ and the Conv4, Conv5 layers in Resnet38 to generate the feature matrix $C \times H \times W$, which is used by CAM2 to perform the classification task. $C$ denotes the number of channels of the features, and $H$ and $W$ denote the height and width of the feature map. First, the feature map is spread along the feature space, and then the two connected layers are allowed to feature the features separately to map the features to the query and key spaces:

$$Q = W_Q F, K = W_K F \quad (2)$$

where $W_Q \in T^{C' \times C}$ and $W_K \in T^{C' \times C}$ are parameters that can be continuously updated and s is a constant. We then use the softmax of the query and keyword inner product to compute the feature space attention perception matrix $A \in T^{HW \times HW}$ [1]:

$$Atten = Soft \max(Q^T K) \quad (3)$$

*Atten* is a measure of the similarity between features corresponding to different feature spaces. We then use matrix multiplication to process the feature attention space maps of the branching networks $\psi_1$ and $\psi_3$ by the procedure:

$$\hat{M}_{CAM_1} = M_{CAM_1} \downarrow A,$$
$$\hat{M}_{CAM_2} = M_{CAM_2} \downarrow A, \quad (4)$$
$$\hat{M}_{CAM_3} = M_{CAM_3} \downarrow A$$

where $\downarrow$ is the down-sampling and Flattening operation. Secondly, we also need to normalize the feature matrix and suppress the non-maximum activation values to zero so as to highlight the maximum activation values. After the adjustment of the attention perception matrix, the final refined feature matrix is obtained.

*D. Network Training*

For the output of each branch we first compute the classification loss in a supervised manner, which continuously optimizes the prediction for that branch. On each independent branch we use a strategy of optimizing around the multi-label soft margin loss:

$$l_{cls}(x,y) = -\frac{1}{C} * \sum_{i=0}^{L-1} \frac{y(i) * \log((1+\exp(-x(i)))^{-1}) +}{(1-y(i)) * \log(\frac{\exp(-x(i))}{1+\exp(-x(i))})} \quad (5)$$

Among them, $L$ represents the tissue category number. Besides, $y(i)$ and $z(i)$ are the image-level label and predicted logits of category i. The total classification loss can be expressed as:

$$L_{cls} = l_{cls_1} + l_{cls_2} + l_{cls_3} \quad (6)$$

Since the features of branch one and branch three need to maintain feature consistency in the refined image generated under the spatial attention feature map. Therefore, the loss of feature consistency is defined as:

$$L_{cons} = \|CAM_1 - CAM_3\| \quad (7)$$

Second, spatial attention feature maps are also affected by feature consistency loss, so we use feature cross-consistency loss to constrain the feature maps of branch one and branch two in the following procedure:

$$L_{cross} = \|CAM_1 - CAM_3\| + \|CAM_3 - CAM_1\| \quad (8)$$

Therefore, the final total loss is:

$$L = L_{cls} + L_{cons} + L_{cross} \quad (9)$$

IV. EXPERIMENTS

*A. Dataset*

WSSS4LUAD is a dataset based on 67 H&E stained slides collected from Guangdong Provincial People's Hospital (GDPH) and 20 WSIs collected from The Cancer Genome Atlas (TCGA) [27]. The training set consisted of 49 WSIs from GDPH and 14 WSIs from TCGA, with 10091 patches cropped. Among them, the patches containing the tumor, stroma, and normal labels were 6579, 7076, and 1832, respectively. The validation set includes 9 WSIs from GDPH and 3 WSIs from TCGA, totaling 40 cropped patches. It includes nine large patches (~1500~5000*1500~5000) and 31 small patches (~200~500*200~500). The test set consists of 9 WSIs from GDPH and 3 WSIs from TCGA with a total of 80 patches cropped (14 large patches and 66 small patches). WSSS4LUAD provides the background masks of the patches for validation and testing purposes only.

*B. Experiment Settings and Implementation Details*

All experiments were based on an Nvidia RTX 4090 GPU with 24GB of memory. The code was implemented using Pytorch 1.13.1 and Pytorch Lightning 1.5.1. In generating the refined mask, the three branches of the CVFC are set with the same learning rate of 0006 and a decay rate of 0.01 for 100 epochs. We use the Unet framework based on the efficientnet-b3 backbone for exact segmentation, with the learning rate set to 0.001 and epochs of 30. Each time the model is trained, The data needs to be preprocessed beforehand, such as flipping, translating, and other operations. For evaluation, Intersection-Over-Union (IoU), mean IoU (mIoU), and frequency-weighted IoU (fwIoU) are used as metrics [28].

## V. RESULTS

### A. Experimental Results and Analysis in WSSS4LUAD

To compare the performance of the models, we chose HistoSegNet, SEAM, C-CAM, WSSS-Tissue, and OEEM methods to compare the mIoU and fwIoU with CVFC on the WSSS4LUAD dataset [21], [29]–[32]. As shown in TABLE I, , the experimental results showed that the CVFC obtained the best mIoU and fwIoU. Secondly, the OEEM method could also accurately segment Tumor and Stroma tissues on pathology images, but not the Normal tissue portion. The other methods performed better in segmenting Tumor and Normal tissues but poorer segmentation on Stroma. We believe that the attention-based cross-view feature consistency principle in the co-training paradigm can constrain CVFC to obtain better segmentation performance. The poor performance of HistoSegNet, SEAM, C-CAM, and OEEM may be because only the CAM was adjusted, and no attentional feature adjustment was made using the original input feature features, resulting in poor generalization performance of the models[21], [29]–[32].

TABLE I. PERFORMANCE COMPARISON ON THE WSSS4LUAD DATASET. WE BOLD THE HIGHEST METHOD AND UNDERLINE THE SECOND HIGHEST METHOD.

| Method | Tumor IoU | Stroma IoU | Normal IoU | mIoU | fwIoU |
|---|---|---|---|---|---|
| HistoSegNet | 0.6521 | 0.3975 | 0.4610 | 0.5035 | 0.5424 |
| SEAM | 0.6425 | 0.4019 | 0.6981 | 0.5808 | 0.5459 |
| C-CAM | 0.6140 | 0.2535 | 0.6337 | 0.5004 | 0.4673 |
| WSSS-Tissue | 0.7471 | 0.5804 | 0.3327 | 0.5534 | 0.6667 |
| OEEM | 0.7597 | 0.6104 | 0.7462 | 0.7054 | 0.6983 |
| CVFC | 0.7846 | 0.5907 | 0.7613 | 0.7122 | 0.7018 |

Second, in Fig. 2, we draw a visual comparison of the test image with the Ground Truth, and we find that the CVFC blurs the segmentation only on some of the Stroma tissues, and the segmentation is more overlapping on the background and Tumor tissues.

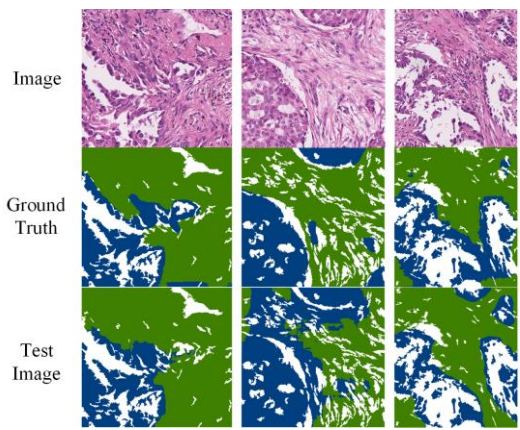

Fig. 2. Comparison of the effectiveness of CVFC for segmentation on test images.

### B. Ablation Study

To validate the effectiveness of our proposed model, we use ablation learning to validate the effectiveness of the attention-based cross-view consistency principle. We use Resnet50, Resnet38 as the models for generating the CAM. We train Resnet50, Resnet38 and CVFC individually to generate the CAM and finally obtain the mask. After using the training data and mask to train the model a supervised model, we again compare the performance of the model using mIoU and fwIoU. As shown in TABLE II, we find that the approach using the attention-based principle of cross-view consistency leads to better segmentation. It may be because the Resnet network alone cannot handle the segmentation boundaries and is not able to accurately convert CAM to Mask in a complex segmentation environment.

TABLE II. ALIDATING THE EFFECTIVENESS OF MODEL SEGMENTATION BASED ON THE EFFECTIVENESS OF CROSS-VIEW CONSISTENCY METHODS.

| Model | Resnet38 | Resnet50 | CVFC |
|---|---|---|---|
| mIoU | 0.6218 | 0.5044 | 0.7122 |
| fwIoU | 0.6473 | 0.5636 | 0.7018 |

### C. Discuss

This paper proposes an end-to-end pseudo-label generation framework based on cross-view feature consistency based on the attention mechanism, which successfully solves the problem of insufficient labels in weakly supervised learning. In experiments, we found that both Resnet38 and Resnet50 have better class activation effects to successfully segment different labels in WSI. However, we found that the class activations obtained by multiple models can be tuned by an attention mechanism. Through the principle of cross-view consistency, the activation maps obtained from different models are constrained, and then critically select classes to activate obvious areas and weaken ambiguous areas. WSSS methods based on CAMs will not be replaced in a short time.

Obviously, there are some shortcomings in the experiment, such as the large training parameters of the model, and the long time for the experiment to generate pseudo-labels. Resnet is a classic model with a large amount of parameters, therefore, we need to increase our computing resources to train larger models. Secondly, if the computing resources are limited, the model can be optimized by using the method of model compression. CVFC is also a data-driven framework. When the amount of data is sufficient, the training effect of the model is better. When the amount of data is insufficient, the model may overfit and be sensitive to abnormal data. Finally, it has to be mentioned that the CVFC model has some inherent defects in some interpretability.

## VI. CONCLUSION

In this paper, we propose an end-to-end pseudo-label generation framework named CVFC based on cross-view feature consistency based on the attention mechanism. Specifically, CVFC is a three-branch federated framework consisting of two resnets38 and one resnet50, where independent branches integrate the feature maps at multiple scales to generate CAMs; in each branch, the size of the CAMs is adjusted by down-sampling and unfolding to adjust the size of the CAM; the intermediate branch projects the feature matrix to the query and key feature space, and generates the feature space-aware matrix by connecting the layers and inner product to adjust and refine the CAM in each branch; and finally optimizes the parameters of the CVFC by the loss of feature consistency and the loss of feature crossover in the co-training mode. An IoU of 0.7122 and a fwIoU of 0.7018 are obtained on the WSSS4LUAD dataset, which outperforms HistoSegNet, SEAM, C-CAM, WSSS-Tissue, and OEEM, respectively.

ACKNOWLEDGMENT

This work was supported by NSFC Grants U19A2067; National Key R&D Program of China 2022YFC3400400; Top 10 Technical Key Project in Hunan Province 2023GK1010, Key Technologies R&D Program of Guangdong Province (2023B1111030004 to FFH). The Funds of State Key Laboratory of Chemo/Biosensing and Chemometrics, the National Supercomputing Center in Changsha (http://nscc.hnu.edu.cn/), and Peng Cheng Lab.